\documentclass[letterpaper, 10 pt, conference]{ieeeconf}  

\IEEEoverridecommandlockouts                              
\usepackage{amsmath}
\usepackage{graphicx}   
\usepackage{placeins}   
\usepackage{dblfloatfix} 
\usepackage{booktabs}
\usepackage{graphicx}
\usepackage{xcolor}
\usepackage{soul}
\newcommand{\mycomment}[1]{}

\title{\LARGE \bf
Design and Modeling of a Single-Port {\color{black}Three}-Arm Robotic Tool for Minimally Invasive Neurosurgery}

\author{Nazia H. Dana$^{*}$, Harith S. Gallage$^{*}$, Ismail A. Auta, Dhanvi Yuvaraj, and Ronghuai Qi, \IEEEmembership{Member, IEEE} %
\thanks{$^{*}$Authors contributed equally to this work.}%
\thanks{This work was supported in part by the Faculty Top Tier Doctoral Graduate Research Assistantship (TTDGRA) Award from the University of Nevada, Las Vegas. \emph{(Corresponding author: Ronghuai Qi.)}}
\thanks{The authors are with the Robotics and Healthcare Systems (RoboHS) Laboratory, Department of Mechanical Engineering, University of Nevada, Las Vegas (UNLV), Las Vegas, NV, USA (e-mail: danan3@unlv.nevada.edu; gallage@unlv.nevada.edu; auta@unlv.nevada.edu; yuvaraj@unlv.nevada.edu; ronghuai.qi@unlv.edu).}
}

\begin{document}

\maketitle
\thispagestyle{empty}
\pagestyle{empty}

\begin{abstract}
Surgical robots require highly dexterous and compact robotic systems capable of operating effectively within confined anatomical spaces. However, due to limited access provided by a single incision, the miniaturization and maneuverability of these robots still need to be improved. In this paper, we propose the design and modeling of a single-port {\color{black}three}-arm robotic tool containing one major cannula (7.14 mm outer diameter (OD)) and three steerable minor cannulas (1.93 mm {\color{black}OD}). By integrating the proposed 12 degrees-of-freedom (DoFs) steerable robotic tool with a 7-DoF robotic arm, this robotic system can potentially achieve multi-arm manipulation capability. We present the design of the steerable robotic tool consisting of tendon-driven joints controlled by a compact actuation system, derive the kinematic model, and validate both the static and kinematic models through experiments.

The performance is evaluated with {\color{black}the root mean square error (RMSE) and mean absolute error (MAE)} computed between the experimental data and the kinematic model.

\end{abstract}

\section{INTRODUCTION}

\begin{figure}[t]
    \centering
    \vspace{0.2cm}
    \includegraphics[width=0.95\columnwidth]{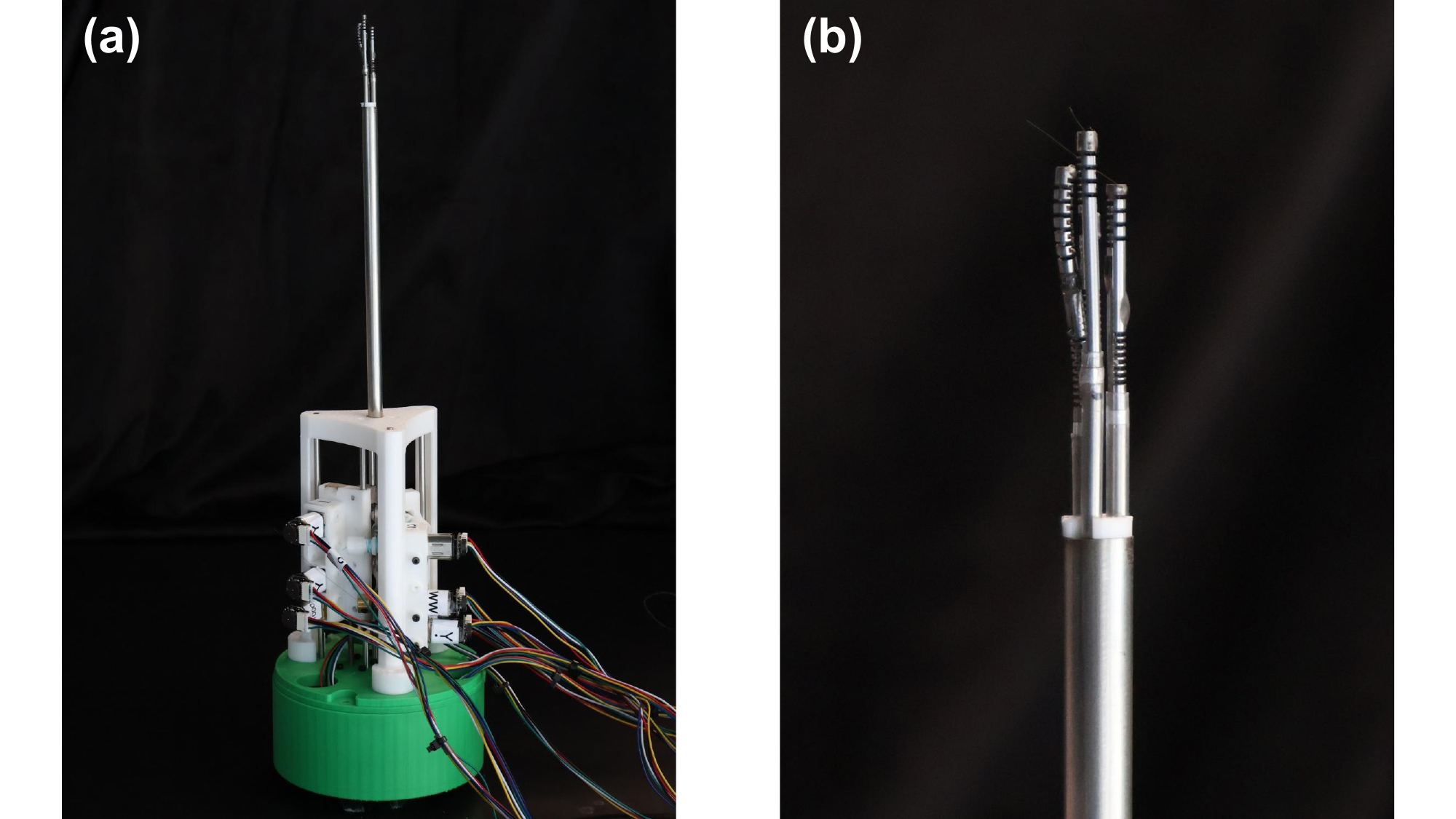}
    \caption{ Single-port {\color{black}three}-arm robotic tool for minimally invasive neurosurgery. (a) {\color{black}E}ntire system and (b) close-up view of the minor cannulas.}
    \label{fig:intro}
    \vspace{-0.7cm}
\end{figure}

 Single-port (SP) robotic platforms have been adopted across a wide range of surgical procedures, providing a minimally invasive approach that can {\color{black}improve surgeon ergonomics and cosmetic outcomes while reducing postoperative complications}\mycomment{improve surgeon ergonomics, cosmetic appearance{\color{black},} and reduce postoperative complications} \cite{biasatti}. Instruments are deployed through an endoscope’s working channels in robot-assisted minimally invasive surgery and multi-channel SP systems enable multi-instrument operation through a single incision to reduce tissue damage \cite{tada2024robotic}. Fig.~\ref{fig:intro} shows the proposed single-port {\color{black}three}-arm robotic tool, including the overall system and a close-up view of the minor cannulas. The system is designed to enable multi-instrument manipulation within the constrained workspace of the system while maintaining a compact single-port architecture.

Clinical data and initial clinical experience with SP robotic platforms have been reported in the literature \cite{biasatti,bianco2022robotic}.
 The reported results suggest that a broad range of major robotic urologic procedures are technically feasible using these systems. One widely used platform is the da Vinci SP system (Intuitive Surgical Inc., Sunnyvale, CA), which has been successfully used in minimally invasive surgery (MIS). The system integrates a stereoscopic binocular camera and three flexible instruments, all occupied in a $25\,\mathrm{ mm}$ diameter cannula \cite{park}. Rox et al. \cite{rox2020mechatronic} introduced a two-arm concentric tube robot for neuroendoscopy, demonstrating coordinated tool motion with limited workspace flexibility due to the passive nature of the outer endoscope structure. This system highlights the advantages of distal bending for manipulation, while also exposing challenges related to mechanical coupling and scalability. {\color{black} \mycomment{The 2-DoF tendon-driven robotic neuroendoscope tool presented in \cite{yamamoto2024preclinical}, can form S-shaped configurations while meeting a $2 \,\mathrm{ mm}$ diameter constraint}}{In our prior work \cite{yamamoto2024preclinical}, we developed a novel 2-DoF tendon-driven robotic neuroendoscope tool that can form S-shaped configurations while meeting a $2 \,\mathrm{ mm}$ diameter constraint.} This design uses micromachined asymmetric notches in a nitinol tube to achieve further bending and reach areas that rigid tools cannot touch.
\mycomment{Additionally, in \cite{chitalia}, the proposed design exhibited inter-joint coupling, causing the actuation of one joint to influence the motion of another.} {\color{black}Additionally, the designs proposed in \cite{chitalia}, exhibited inter-joint coupling, where the actuation of one joint influenced the motion of another.} The present work addresses this {\color{black}issue} by minimizing inter-joint coupling and improving independent joint control.

 A key limitation of these studies is that bending is highly sensitive to tendon displacement and joint statics, which can make control more susceptible to noise and reduce accuracy. The rigidity of most current surgical tools limits distal dexterity, triangulation{\color{black},} and access to the full workspace. To address these limitations, continuum and tendon-driven robotic tools have been developed to improve distal articulation while maintaining compatibility with rigid neuroendoscopes.

\begin{figure*}[t!]
    \centering
    \includegraphics[width=\textwidth]{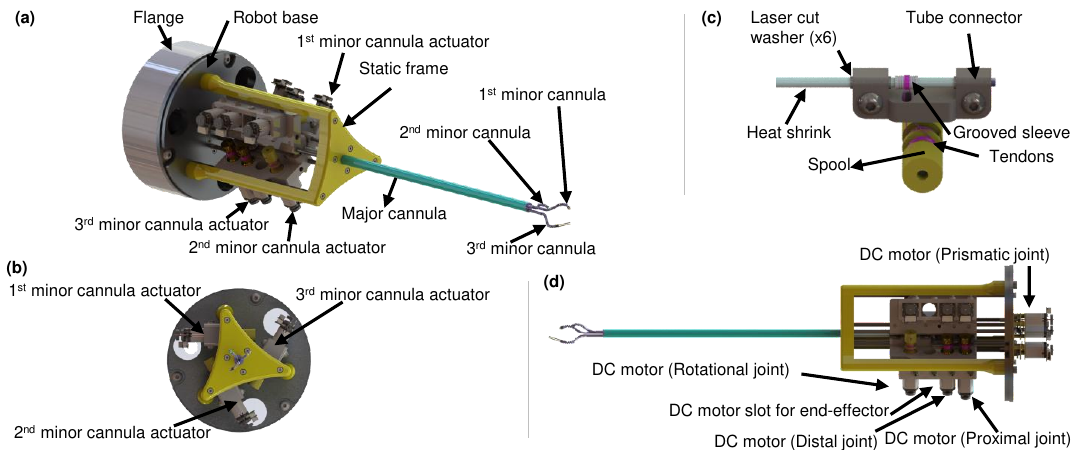}
    \caption{Schematic of the compact actuation system (CAS), (a) an exploded view of CAS with three steerable cannulas, (b) top view of the CAS, (c) redesigned spool, and (d) an exploded view of minor cannula actuator.}
    \label{fig:Cas}
    \vspace{-0.7cm}
\end{figure*}

 Many platforms still focus on single-channel tool deployment, which limits coordinated multi-instrument manipulation \cite{tada2024robotic}. To address these limitations, the proposed steerable multi-cannula system enables coordinated {\color{black}three}-arm functionality within a single neurosurgical access port. Moreover, neurosurgical applications impose substantially stricter diameter and workspace constraints, requiring miniaturized mechanisms and highly compact kinematic architectures to ensure safe and effective operation within confined intracranial environments \cite{butler2012robotic}. Additionally, systematic reviews on continuum and soft robots for MIS highlight the advantages of flexible, compliant manipulators in confined anatomical spaces, while noting that most designs focus on single-instrument systems \cite{iqbal2025continuum}. 

In our previous {\color{black}work} \cite{qi2025development}, we developed a SP dual-arm robotically steerable endoscope for neurosurgical applications that integrates two tendon-driven cannulas capable of distal bending while maintaining a compact access diameter. 
Building on \cite{qi2025development}, this work introduces the novel design of a SP multi channel major cannula (OD: $7.14 \,\mathrm{mm}$ and ID: $6.54 \,\mathrm{ mm}$) and three steerable minor cannulas (OD: $1.93 \,\mathrm{ mm}$ and ID: $1.49 \,\mathrm{ mm}$). Compared to the systems discussed above, the proposed device can be introduced through a smaller single incision while providing a larger reachable workspace, improved dexterity{\color{black},} and reduced inter-joint coupling than current state-of-the-art approaches.

The main contributions of this study are summarized as follows:
\begin{itemize}[]
  \item Design and modeling of a 12-DoF single-port, {\color{black}three-}arm robotic tool that consists of prismatic, rotational, proximal bending{\color{black},} and distal bending joints.
  \item Development of a modular, compact actuation architecture for the steerable joints of three minor cannulas.

\item {\color{black}Derivation and experimental validation of the kinematic model for the proposed robotic system.}

\end{itemize}

The paper is organized as follows: Section II highlights the system design. Section III presents the robot kinematic modeling, while Section IV discusses experimental results and evaluation of performance. Finally, Section V summarizes the conclusions and future work.

\section{System Design}

\begin{figure}[t]
    \centering
    \includegraphics[width=1\linewidth]{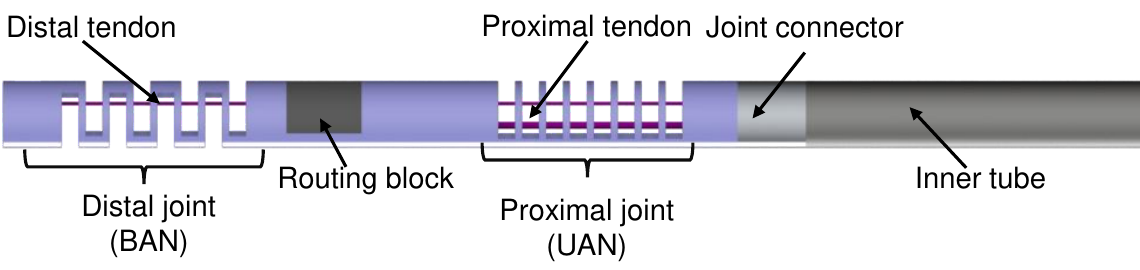}
    \caption{Schematic of the robotically steerable minor cannula comprising of {\color{black}a} proximal and {\color{black}a} distal joint. }
    \label{fig:cannula}
    \vspace{-0.8cm}
\end{figure}

In this study, three minor cannulas are placed inside one major cannula{\color{black},} as shown in Fig.~\ref{fig:Cas}(a). 
Each minor cannula is steerable and has two independently actuated bending joints: a proximal joint and a distal joint. This configuration is demonstrated in Fig.~\ref{fig:cannula}.

The bending functionality of the proximal and distal joints is achieved through laser micromachining of specific notch patterns along the cannula tube{\color{black},} as {\color{black}illustrated} in  Fig.~\ref{fig:cannula}. These notches locally reduce the stiffness of the tube in a controlled manner, allowing it to bend preferentially in a defined direction when a tendon is tensioned. The proximal joint utilizes a unidirectional asymmetric notch (UAN) pattern, which enables controlled bending in one direction. The distal joint employs a bidirectional asymmetric notch (BAN) pattern, enabling bending in either direction within the same plane. The notch depth and spacing in both the proximal and distal sections are designed as 0.5 mm and 0.2 mm, respectively, while the depths of cut are 1.68 mm in the proximal section and 1.6 mm in the distal section. Among the three minor cannulas, two share an identical configuration with 8 notches in both the proximal and distal bending sections, resulting in proximal and distal joint lengths of 5.6 mm each. The third cannula has a slightly different configuration, consisting of 10 notches in the proximal section and 12 notches in the distal section, yielding proximal and distal joint lengths of 7 mm and 8.4 mm, respectively. Both bending joints are actuated by tendons that are wound around and fixed to spools using {\color{black}a} pull{\color{black}-}and{\color{black}-}release mechanism.

The minor cannulas provide axial translation via the prismatic joint, allowing them to be advanced beyond or retracted back into the major cannula. The joint provides a linear travel of 40 mm.

In \cite{qi2025development}, tendon slack was observed during the actuation of rotational joint of the minor cannula. To address this issue, two improvements were introduced. First, the tendon was wound around the grooved sleeve using one slot in the tube connector to ensure better guidance as shown in Fig.~\ref{fig:Cas}(c). Second, the brass spool was replaced with a 3D-printed Polylactic Acid (PLA) spool designed with two dedicated slots for secure tendon attachment. The redesigned spool includes a middle wall on the cylindrical surface, allowing the two tendon segments to be wound separately after passing through the grooved sleeve. Together, these design modifications significantly reduce tendon slack and improve the precision and accuracy of rotational motion.

\begin{figure}[t]
    \centering
    \includegraphics[width=1\columnwidth]{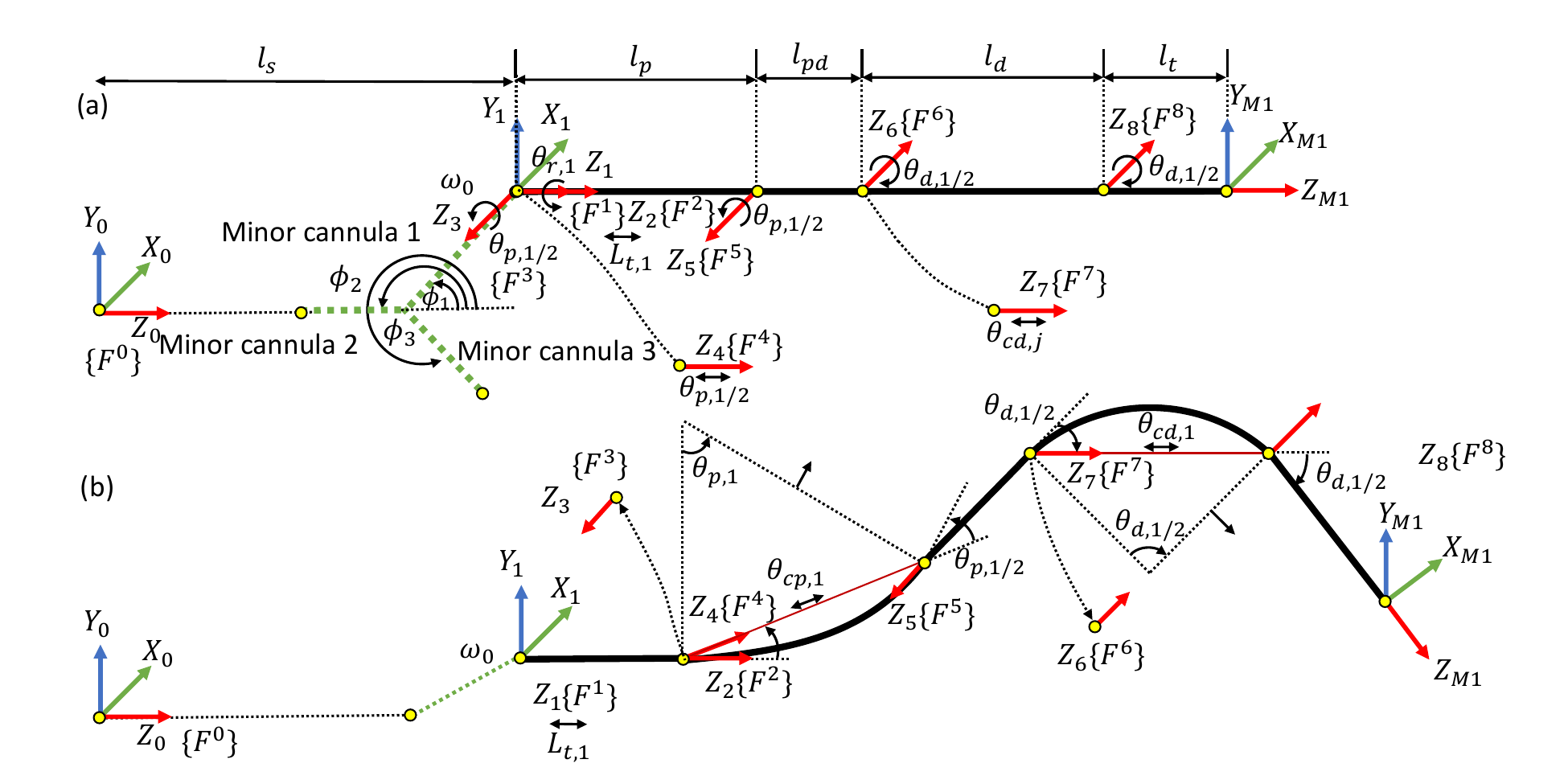}
    \caption{Schematic of the minor cannulas with coordinate frames attached to the joints. (a) Unactuated state and (b) actuated state.}
    \label{fig:frames}
    \vspace{-0.7cm}
\end{figure}

The compact actuation system integrates 12 joint actuators, with the bending joints driven through a tendon-based actuation mechanism. DC gearmotors (HPCB 6V dual-shaft, Pololu Corp., NV) of 380:1 gear ratio are used for distal and rotational joint{\color{black}s}, {\color{black}a} 1000:1 gear ratio motor is used for proximal bending joint{\color{black},} and 100:1 lead screw gear motor is used for actuating the prismatic joint,  as illustrated in Figs.~\ref{fig:Cas}(b) and~\ref{fig:Cas}(d).

 Each minor cannula provides 4-DoF per arm, comprising prismatic insertion, axial rotation, and independent proximal and distal bending. Additionally, each  actuator module of minor cannula includes an extra slot designed to accommodate a tendon-driven actuator (similar to those used for the proximal or distal joints) for operating an end-effector, such as a grasper or scissors, although this feature is not implemented in the present work. By integrating a 7-DoF robotic arm with a wireless joystick interface, the overall system has the potential to enable intuitive teleoperated MIS procedures.

\section{Kinematic Modeling}

The four actuated joints of each minor cannula are: (i) one prismatic joint for linear advancement, (ii) one axial revolute joint for rotation, and (iii) two bending joints corresponding to the proximal and distal steerable sections. All four joints are explicitly incorporated into the forward kinematics using the Product of Exponentials (POE) method \cite{murray}. The geometric parameters used in the kinematic model follow the notation illustrated in Fig.~\ref{fig:frames}. The parameters $l_s$, $l_p$, $l_{pd}$, $l_d$, and $l_{t}$ represent the distance between the base and the minor cannula, the proximal bending section length, the spacer length between the proximal and distal joints, the distal bending section length, and the straight tip segment length, respectively. The prismatic insertion and rotational angle are denoted by $L_t$ and $\theta_{r}$, while the proximal and distal joint angles are represented by $\theta_{p}$ and $\theta_{d}$. 
Fig.~\ref{fig:frames}(a) shows the unactuated configuration of one minor cannula when all joint variables are zero, while Fig.~\ref{fig:frames}(b) shows the actuated configuration when nonzero bending, prismatic{\color{black},} and rotational inputs are applied. In this formulation, the prismatic joint responsible for linear insertion of the minor cannula is explicitly included in the POE representation and contributes directly to the overall forward kinematics of the system.

\begin{figure}[t]
    \centering
                                                                         \includegraphics[width=0.8\linewidth]{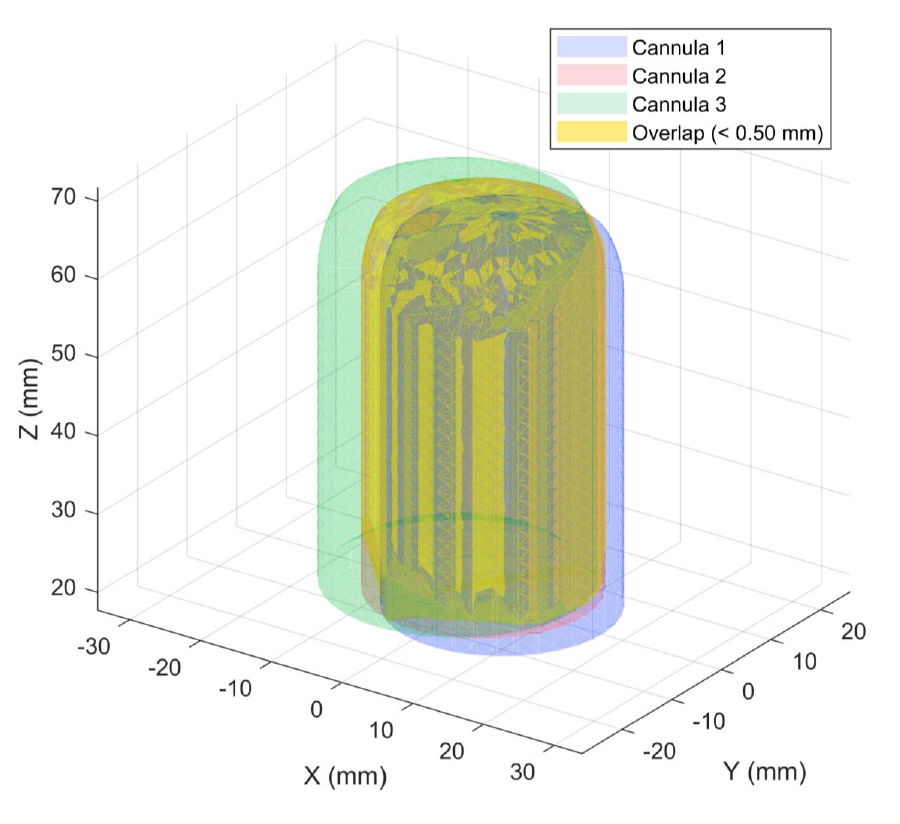}
    \vspace{-0.2cm}
    \caption{3D workspace of the three minor cannulas showing the overlap region.}
    \label{fig:workspace}
    \vspace{-0.8cm}
\end{figure}

The base of each minor cannula is offset from the center of the major cannula by a distance $w_0$, which represents the radial distance between the axis of the major cannula and the base axis of each minor cannula. $\phi_0$ denotes the initial angular offset and the base of the $j$-th minor cannula is located at angular position $(\phi_j)$ about the major cannula’s central axis, defined as:

\begin{equation}
\phi_j = \phi_0 + (j-1)\frac{2\pi}{n}, 
\qquad j \in \{1,2,\dots,n\}
\label{eq:phi}
\end{equation}
where $n$ denotes the number of minor cannulas. The spatial rotation $(R_j)$ and translation $(p_j)$ of {\color{black}a}rm-$j$ are given by:
\begin{equation}
R_j = R_z(\phi_j), \qquad
p_j =
\begin{bmatrix}
w_0 \cos\phi_j \\
w_0 \sin\phi_j \\
l_s
\end{bmatrix}.
\label{eq:Rj_pj}
\end{equation}

The kinematic structure of each minor cannula consists of one prismatic joint for insertion, one axial rotational joint and two bending sections {\color{black}each} modeled as revolute–prismatic–revolute (RPR) mechanism. {\color{black}Accordingly, the complete kinematic chain contains eight joints in total, and the POE formulation requires eight twists to represent the forward kinematics of each arm.} \mycomment{the complete kinematic chain contains eight joints in total. {\color{black}Therefore}, the POE formulation requires eight twists to represent the forward kinematics of each arm. Therefore, 8 twists are required in the POE formulation to represent the forward kinematics of each arm.}

The twist of a rotational and prismatic joint are respectively given by: 
\begin{equation}
\xi =
\begin{bmatrix}
- \omega \times q \\
\omega
\end{bmatrix},
\qquad
\xi =
\begin{bmatrix}
v \\
0 
\end{bmatrix}.
\label{eq:twist_def}
\end{equation}
The first two joint twists of {\color{black}a}rm-$j$ are:
\begin{equation}
\xi_{1,j} =
\begin{bmatrix}
0 \\ 0 \\ 1 \\ 0 \\ 0 \\ 0
\end{bmatrix},
\qquad
\xi_{2,j} =
\begin{bmatrix}
w_0 \sin\phi_j \\
- w_0 \cos\phi_j \\
0 \\
0 \\ 0 \\ 1
\end{bmatrix}.
\label{eq:first_twists}
\end{equation}

The remaining twists $\xi_{i,j}$ for $i=3,\dots,8$, shown in Eq. \eqref{eq:7} are obtained similarly by selecting the appropriate axis direction (via $R_j$) and a point $q_{i,j}$ on the axis at the corresponding {\color{black}$Z$}-location, then applying $v=-\omega\times q$ for rotational joints and $v$ for prismatic joints.

 \begin{figure}[t]
    \centering
    \includegraphics[width=0.7\linewidth]{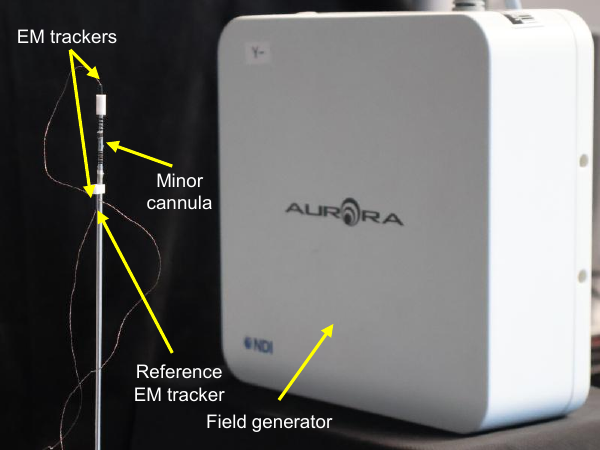}
    \caption{Experimental setup.}
    \label{fig:experimental_setup}
    \vspace{-0.7cm}
\end{figure}

Here, the twist $\xi_{1,j}$ corresponds to the prismatic insertion joint and $\xi_{2,j}$ corresponds to the axial rotation joint. The twists $\xi_{3,j}$, $\xi_{4,j}$, and $\xi_{5,j}$ correspond to the proximal RPR joint sequence, while the twists $\xi_{6,j}$, $\xi_{7,j}$ and $\xi_{8,j}$ correspond to the distal RPR joint sequence.

\begin{equation}
\small 
\begin{aligned}
\xi_{3,j} &=
\begin{bmatrix}
l_s\sin\phi_j\\
- l_s\cos\phi_j\\
0\\
-\cos\phi_j\\
-\sin\phi_j\\
0
\end{bmatrix},
\qquad
\xi_{4,j}=
\begin{bmatrix}
0\\0\\1\\0\\0\\0
\end{bmatrix},
\\[6pt]
\xi_{5,j} &=
\begin{bmatrix}
(l_s+l_p)\sin\phi_j\\
-(l_s+l_p)\cos\phi_j\\
0\\
-\cos\phi_j\\
-\sin\phi_j\\
0
\end{bmatrix},
\quad\mspace{-18mu} 
\xi_{6,j}=
\begin{bmatrix}
-(l_s+l_p+l_{pd})\sin\phi_j\\
(l_s+l_p+l_{pd})\cos\phi_j\\
0\\
\cos\phi_j\\
\sin\phi_j\\
0
\end{bmatrix},
\\[6pt]
\xi_{7,j} &=
\begin{bmatrix}
0\\0\\1\\0\\0\\0
\end{bmatrix},
\qquad
\xi_{8,j}=
\begin{bmatrix}
(l_s+l_p+l_{pd}+l_d)\sin\phi_j\\
-(l_s+l_p+l_{pd}+l_d)\cos\phi_j\\
0\\
\cos\phi_j\\
\sin\phi_j\\
0
\end{bmatrix}.
\end{aligned}
\label{eq:7}
\end{equation}
\vspace{-0.2cm}
\normalsize

Let the total straight length $(L_{\mathrm{tot}})$ be,
\begin{equation}
L_{\mathrm{tot}} = l_s + l_p + l_{pd} + l_d + l_{t} 
\label{eq:Ltot_compact}.
\end{equation}

The home configuration of {\color{black}a}rm-$j$ is given by:
\begin{equation}
g_{0m_j}(0)=
\begin{bmatrix}
I_{3\times 3} &
\begin{bmatrix}
w_0\cos\phi_j\\
w_0\sin\phi_j\\
L_{\mathrm{tot}}
\end{bmatrix}\\
0\ 0\ 0 & 1
\end{bmatrix}.
\label{eq:g0mj_compact}
\end{equation}

For the three-arm case $({\color{black}n=3, \phi_0=\frac{\pi}{3}})$, substituting 
$\phi_1=\frac{\pi}{3},\, \phi_2=\pi,\, \phi_3=\frac{5\pi}{3}$ yields:
\begin{equation}
\begin{aligned}
g_{0m_1}(0) &=
\begin{bmatrix}
I_{3\times 3} &
\begin{bmatrix}
\frac{1}{2}w_0\\
\frac{\sqrt{3}}{2}w_0\\
L_{\mathrm{tot}}
\end{bmatrix}\\
0\ 0\ 0 & 1
\end{bmatrix},\\[10pt]
g_{0m_2}(0) &=
\begin{bmatrix}
I_{3\times 3} &
\begin{bmatrix}
- w_0\\
0\\
L_{\mathrm{tot}}
\end{bmatrix}\\
0\ 0\ 0 & 1
\end{bmatrix},\\[10pt]
g_{0m_3}(0) &=
\begin{bmatrix}
I_{3\times 3} &
\begin{bmatrix}
\frac{1}{2}w_0\\
-\frac{\sqrt{3}}{2}w_0\\
L_{\mathrm{tot}}
\end{bmatrix}\\
0\ 0\ 0 & 1
\end{bmatrix}.
\end{aligned}
\label{eq:g0mj_three_compact}
\end{equation}
%

Eq.~(\ref{eq:g0mj_three_compact}) provides the home configuration of each arm.
\vspace{2mm}

\begin{figure}[t]
    \centering
    \includegraphics[width=0.8\columnwidth]{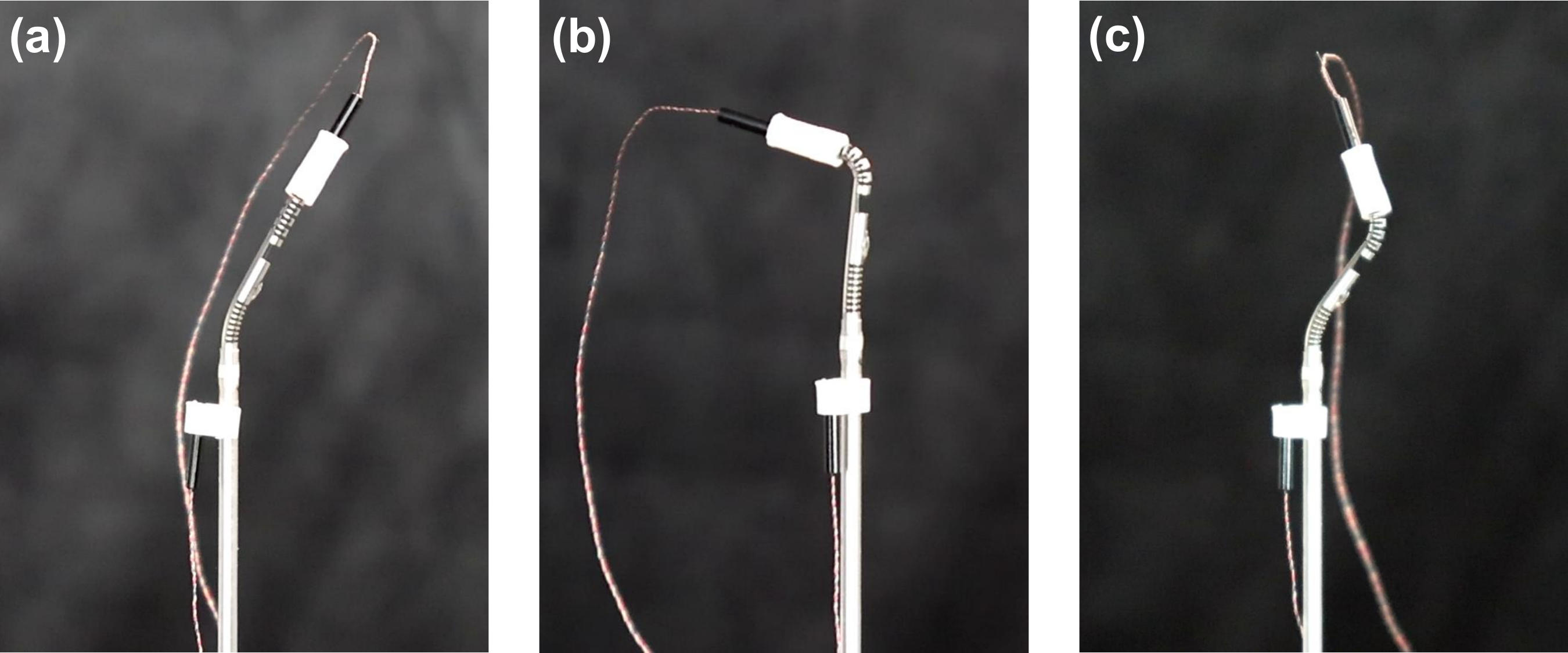}
    \caption{Actuation of bending joints, (a) proximal, (b) distal {\color{black},} and (c) both proximal and distal.}
    \label{fig:actuation_joints}
    \vspace{-0.7cm}
\end{figure}

The workspace of the three minor cannulas obtained from the forward kinematics model is shown in Fig.~\ref{fig:workspace}. The workspace is generated by providing joint inputs of 40~mm for the prismatic insertion, 30$^\circ$ for the proximal bending, 75$^\circ$ for the distal bending{\color{black},} and 360$^\circ$ for the axial rotation. As shown in the Fig. 5, the workspaces of the three cannulas partially overlap, creating a common region that can be used for coordinated manipulation. Due to the longer proximal and distal bending sections, the third cannula achieves a slightly larger lateral workspace compared to the other two cannulas. The overall reachable workspace spans approximately 34.77~mm, 35.47~mm, and 53.99~mm along the $X$, $Y$, and $Z$ directions, respectively.

\section{Results \& Discussion}

\begin{figure*}[t]
    \centering
    \includegraphics[
        width=0.95\textwidth]{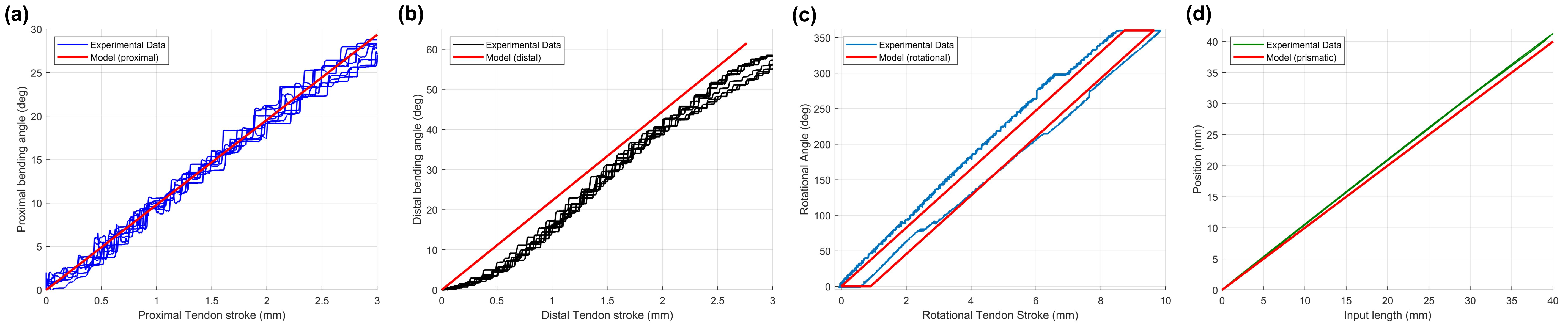}
    \caption{Experimental validation of joint static models of (a) proximal joint, (b) distal joint, (c) rotational joint{\color{black},} and (d) prismatic joint. }
    \label{fig:Joint}
    \vspace{-0.5cm}
\end{figure*}

\begin{figure}[!b]
    \centering
    \includegraphics[width=1\columnwidth]{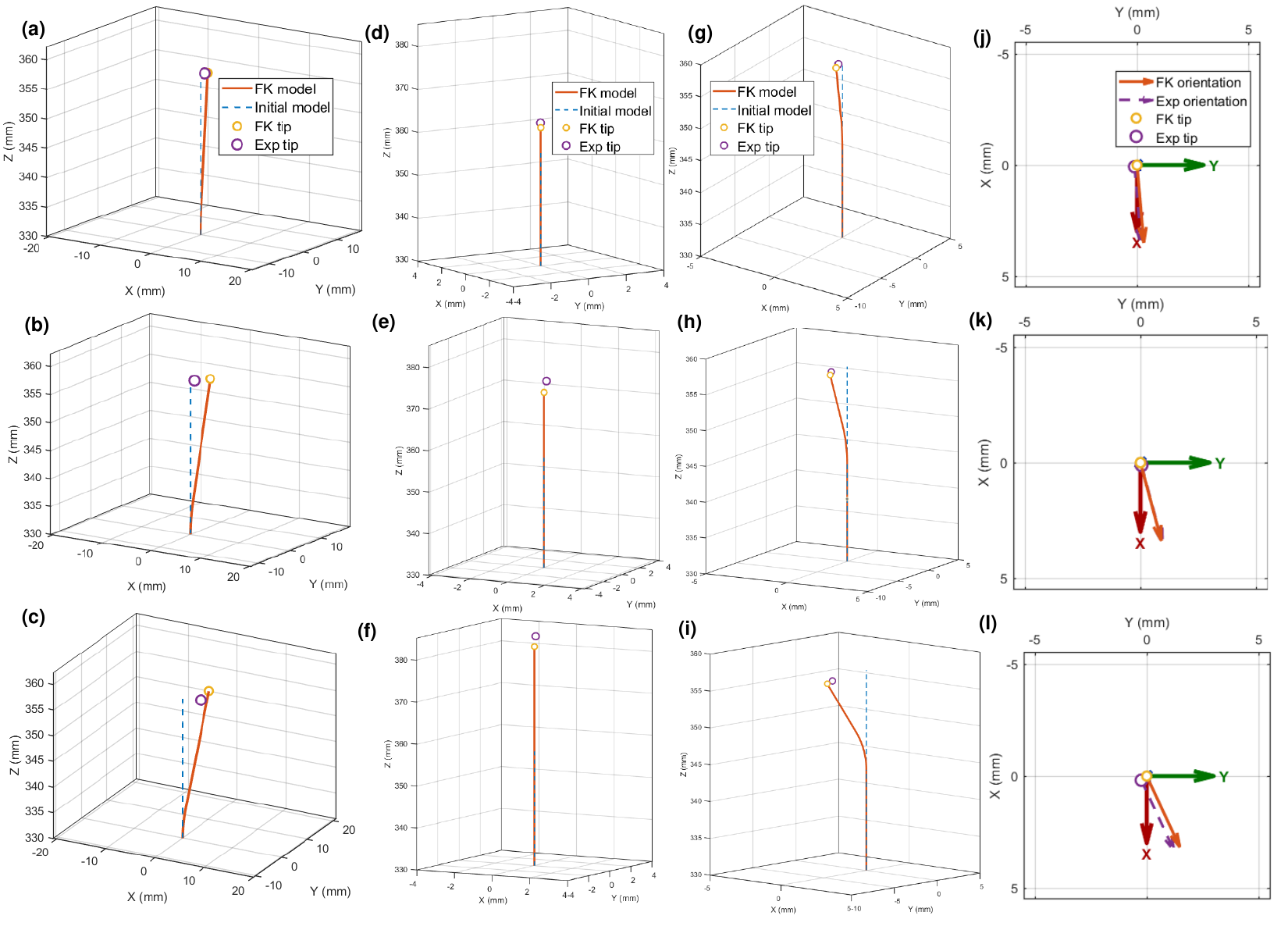}
    \caption{Kinematic model validation of the (a)-(c) proximal, (g)-(i) distal, (j)-(l) rotational joints at $5^\circ$, $15^\circ${\color{black},} and $25^\circ$, respectively, and (d)-(f) prismatic joint at 5\,mm, 15\,mm{\color{black},} and 25\,mm insertion.}
    \label{fig:fk}
\end{figure}

Two sets of experiments were conducted to validate the proposed models. First, each joint was actuated independently to verify the corresponding static and kinematic models.
{\color{black}\mycomment{Experiments were conducted to independently evaluate each joint by comparing the experimental results with the static and kinematic models.}} In addition, combined actuation involving all joints was performed to assess the simultaneous operation of a single arm and validate the kinematic model. We used a minor cannula with 8 proximal and 8 distal notches for the experiments.  
To {\color{black}set up} the experiment (see Fig.~\ref{fig:experimental_setup}), a 6-DoF electromagnetic (EM) tracking system (Aurora, Northern Digital Inc., {\color{black}Ontario, Canada})  was used with two sensors to ensure precise measurement of the cannula tip motion.

One sensor served as a fixed reference, while the second tracked the cannula tip motion. The joint static models for the proximal, distal{\color{black},} and rotational joints were adopted from~\cite{qi2025development}. Fig.~\ref{fig:actuation_joints} shows the joint behavior during individual and coordinated actuation of proximal and distal joints. Fig.~\ref{fig:Joint} shows the deflection of joint angles for respective tendon stroke between experimental and static models, while Fig.~\ref{fig:fk} and Fig.~\ref{fig:combined actuation graph} show the tip displacement of experimental and forward kinematics model for the given angles and  the corresponding tip position trajectories over time, respectively. The proposed joint kinematic models were evaluated using {\color{black}RMSE and MAE}.

\mycomment{Fig.~\ref{fig:Joint} shows the measured cannula tip orientation or position of actuating each joint as a function of the commanded input, while Fig.~\ref{fig:combined actuation graph} shows the corresponding tip position trajectories over time. The proposed joint statics and kinematic models were evaluated using the root mean square error (RMSE), and mean absolute error (MAE).
The proposed joint statics and kinematic models were evaluated using the root mean square error (RMSE), and mean absolute error (MAE).
}

\subsection{Verification of Joint Static Model}

In this study, proximal and distal bending experiments were conducted separately with a target deflection of 30$^\circ$ and 60° respectively, as shown in Fig.~\ref{fig:actuation_joints}(a) and \ref{fig:actuation_joints}(b). The experimental measurements closely match the model's output, as illustrated in Fig.~\ref{fig:Joint}(a) and \ref{fig:Joint}(b). Compared with our previous work \cite{qi2025development}, which demonstrated proximal joint actuation up to 15$^\circ$, the present study extends the evaluated actuation range to 30$^\circ$. In addition, Fig.~\ref{fig:actuation_joints}(c) indicates that the proposed system and actuation mechanism support simultaneous actuation of the proximal and distal bending joints, which increases overall dexterity and configuration flexibility.

\begin{figure}[h]
    \centering
    \includegraphics[width=1\columnwidth]{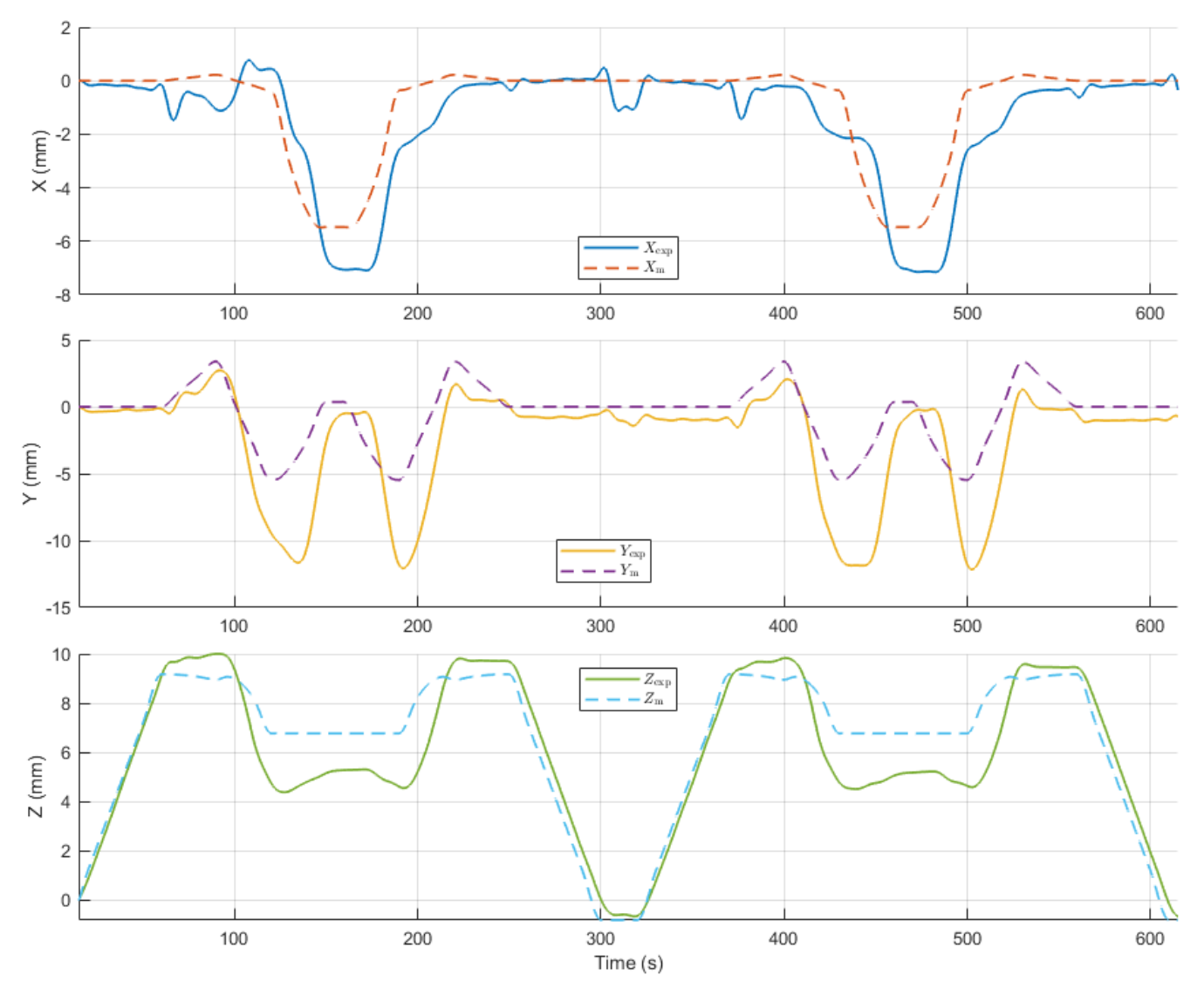}
    \caption{Comparison of experimental and kinematic modeled tip trajectories for simultaneous joint actuation.}
    \vspace{-0.5cm}
    \label{fig:combined actuation graph}
\end{figure}

\mycomment{
{\color{black} In this study, four bending cycles were recorded and compared with the statics models of the proximal and distal joints independently. Proximal and distal bending experiments were conducted separately with a target deflection of 30$^\circ$ and 60° respectively, as shown in Fig.~\ref{fig:actuation_joints}(a) and (b). The experimental measurements of proximal closely match the model's output, with substantial overlap between the two curves as illustrated in Fig.~\ref{fig:Joint}(a). The resulting error metrics are low (RMSE = 1.10$^\circ$, MAE = 0.92$^\circ$), as summarized in Table~\ref{tab:rmse_mae}.
In Fig.~\ref{fig:Joint}(b), the statics model tracks the experimental measurements of distal bending with only a small deviation, demonstrating close alignment between the model results and experimental data. This is reflected in the error metrics (RMSE = 4.38$^\circ$, MAE = 3.66$^\circ$) reported in Table~\ref{tab:rmse_mae}. In addition, Fig.~\ref{fig:actuation_joints}(c) indicates that the proposed system and actuation mechanism support simultaneous actuation of the proximal and distal bending joints, which increases overall dexterity and configuration flexibility.} 
}A full-range rotational actuation experiment was performed in which the joint completed a 360° sweep in one direction and then returned to its initial position, extending our prior design that was limited to 180° \cite{qi2025development}. The rotational joint static model incorporating a 0.9 mm tendon slack compensation term was used for comparison. As shown in Fig.~\ref{fig:Joint} (c), the model captures the overall experimental trend. 

\mycomment{A full-range rotational actuation experiment was performed in which the joint completed a 360° sweep in one direction and then returned to its initial position, extending our prior design that was limited to 180° \cite{qi2025development}. {\color{black}A joint statics model incorporating a 0.9 mm tendon slack compensation term was used for comparison. As shown in Fig.~\ref{fig:Joint} (c), the model captures the overall experimental trend.} 
Although the errors remain relatively high (RMSE = 13.56$^\circ$, MAE = 10.27$^\circ$), particularly across the full bidirectional rotation range. Three actuation cycles were recorded and compared with the statics models of the rotational and prismatic joints independently.}
Prismatic joint experiments were conducted with a 40 mm upward translation followed by a return to the initial position. The corresponding static model is given in Eq. \eqref{eq:prismatic_kinematics_1}, where $p$ denotes the lead-screw pitch (0.5 mm) of the dual-start lead-screw DC gear motor, $x$ is the commanded linear displacement{\color{black},} and $\theta_{\mathrm{pris}}$ is the required motor angle to achieve $x$. As shown in Fig.~\ref{fig:Joint}(d), the experimental measurements closely match the model's output, yielding low errors.

\mycomment{
Prismatic joint experiments were conducted with a 40 mm upward translation followed by a return to the initial position. The corresponding statics model is given in Eq. \eqref{eq:prismatic_kinematics_1}, where $p$ denotes the lead-screw pitch of the dual-start gear motor, $x$ is the commanded linear displacement and $\theta_{\mathrm{pris}}$ is the required motor angle to achieve $x$. As shown in Fig.~\ref{fig:Joint}(d), the experimental measurements closely match the model's output, yielding low errors (RMSE = 0.92 mm, MAE = 0.82 mm). 
}

\begin{equation}
\label{eq:prismatic_kinematics_1}
\theta_{\text{pris}} = \frac{\pi}{p}\,x
\end{equation}

\mycomment{
\begin{figure*}[t]
    \centering
    \includegraphics[
        width=\textwidth]{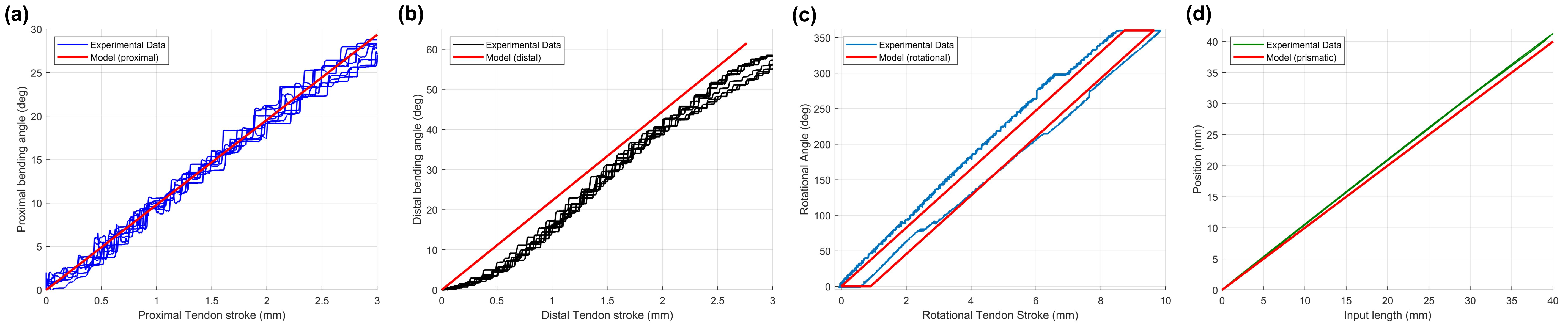}
    \caption{Experimental validation of joint statics models of (a) proximal joint, (b) distal joint, (c) rotational joint and (d) prismatic joint. }
    \label{fig:Joint}
    \vspace{-0.5cm}
\end{figure*}

}

\mycomment{The results indicate that the proximal bending joint shows a low discrepancy between the model and experimental results, demonstrating accurate analytical prediction. The distal joint exhibits moderate deviation.}

 With the new spool design, the rotational joint achieves full $360^\circ$ bidirectional actuation. The prismatic joint maintains small linear error, confirming reliable lead-screw-based translational motion modeling. Overall, the statics results validate that the proposed joint models accurately capture both the angular and translational joint behaviors, with deviations remaining within acceptable ranges for the intended application.
{\mycomment{However, it exhibits comparatively higher error, which is likely attributable to tendon slack.}}

 {\mycomment{   {Forward kinematics model validation for all joints at $5^\circ$, $15^\circ$, and $25^\circ$ or 5 mm, 15 mm, and 25 mm . (a)-(c) Proximal Joint at $5^\circ$, $15^\circ$, and $25^\circ$, (d)-(f) Prismatic Joint at 5 mm, 15 mm, and 25 mm, (g)-(i) Distal Joint at $5^\circ$, $15^\circ$, and $25^\circ$, and (j)-(l) Rotational Joint at $5^\circ$, $15^\circ$, and $25^\circ$,}}

\subsection{Verification of Kinematic Model}
To evaluate the accuracy of the proposed kinematic model under realistic operating conditions, all joints of the minor cannula were actuated independently and simultaneously. 
To ensure consistency across the validation experiments, then  rotational, proximal{\color{black},}  and distal joints were evaluated at $5^\circ$, $15^\circ${\color{black},}  and $25^\circ$, while the prismatic joint was evaluated at corresponding insertion levels of 5 mm, 15 mm{\color{black},} and 25 mm independently, as shown in Fig.~\ref{fig:fk}. In that plot, the $Z$-axis originates at 330 mm, corresponding to the axial distance from the robot base to the proximal joint. The RMSE and MAE values obtained from the experiments are listed in Table ~\ref{tab:rmse_mae} . The RMSE and MAE values are below 1.5 mm and 1 mm, demonstrating the validity and accuracy of the proposed kinematic model.

\begin{figure}[t]
    \centering
    \includegraphics[width=0.9\columnwidth]{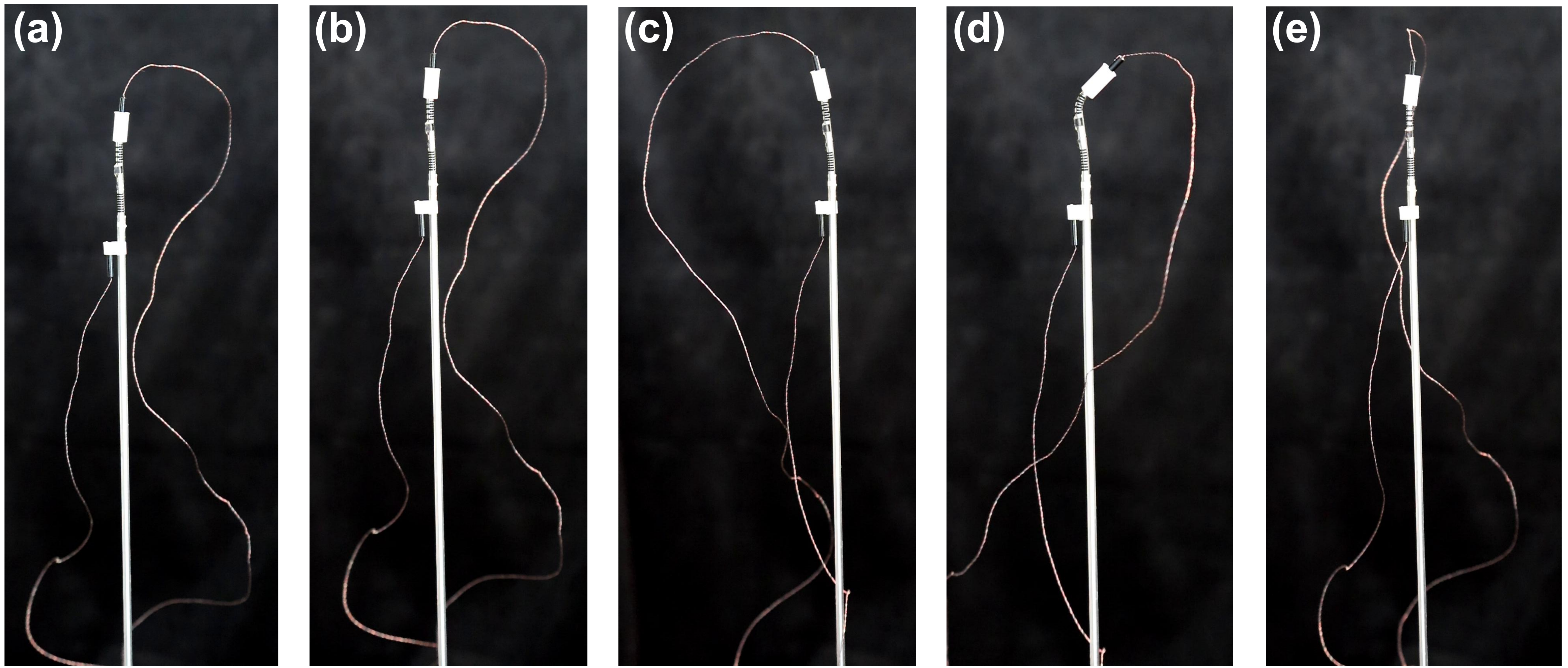}
    \caption{Combined actuation of the joints: (a) initial position, (b) 10 mm prismatic translation, (c) 8$^\circ$ proximal bending, (d) 50$^\circ$ distal bending{\color{black},} and (e) 90$^\circ$ rotation.}
    \label{fig:combined actuation fig}
    \vspace{-0.6cm}
\end{figure}

Fig. \ref{fig:combined actuation graph} compares the experimentally measured cannula-tip trajectories with the model results along the three Cartesian directions. For this experiment, the unactuated tip position shown in Fig.~\ref{fig:frames}(a) was considered as the initial position. 
Overall, the model tracks the motion pattern well throughout the time sequence of 620~s, as shown in Fig. \ref{fig:combined actuation fig}. During the experiment, the tip position was continuously recorded in $X$, $Y${\color{black},} and $Z$ coordinates using an EM tracking system. In parallel, the same joint inputs were applied to the forward kinematics model to compute the predicted tip trajectory. The commanded inputs were applied sequentially: 10 mm prismatic translation, 8$^\circ$ proximal bending, 50$^\circ$ distal bending{\color{black},} and 90$^\circ$ axial rotation. The resulting RMSE values are 1.15~mm, 3.15~mm{\color{black},} and 1.29~mm,  while the corresponding MAE values are 0.84~mm, 2.19~mm{\color{black},} and 1.05~mm along the $X$, $Y${\color{black},} and $Z$ axes, respectively, shown in Table~\ref{tab:rmse_mae}. The $Z$-axis shows the closest agreement, particularly during large elevation changes, while small deviations in the $X$ and $Y$ directions are more apparent during rapid transitions and peak excursions. These discrepancies are likely due to mechanical compliance, tendon friction, minor backlash and{\color{black},} other effects that are not explicitly captured by the ideal kinematic formulation.

    {\mycomment{The prismatic insertion, proximal bending, distal bending and axial rotation were driven together according to a predefined input sequence over a total duration of 620~s, as shown in Fig. \ref{fig:combined actuation fig}. This coordinated actuation was designed to generate three-dimensional motion of the cannula tip.} }

Overall, Cartesian errors remain within a few millimeters, with the largest deviation occurring along the $Y$-axis, supporting the effectiveness of the proposed joint models in capturing both angular and translational behaviors of the robotic system.

\begin{table}[t]
\centering
\caption{Performance Evaluation of Kinematics Models}
\label{tab:rmse_mae}
\begin{tabular}{|l|c|c|}
\hline
Experiment & RMSE {\color{black}(mm)} & MAE {\color{black}(mm)} \\
\hline
Proximal    & 1.43  & 0.99  \\
Distal      & 1.09  & 0.99  \\
Rotational  & 0.22 & 0.21 \\
Prismatic         & 0.92  & 0.82  \\
Combined X        & 1.15  & 0.84 \\
Combined Y        & 3.15  & 2.19 \\
Combined Z        & 1.29  & 1.05 \\
\hline
\end{tabular}
\vspace{-0.7cm}
\end{table}

\section{Conclusions \& Future Work}
The proposed robotic tool consists of three 4-DoF minor cannulas (1.93 mm OD) that can be integrated within 7.14 mm OD, making the system smaller than existing state-of-the-art single-port MIS robots. In this study, the steerable minor cannula achieves a full $\pm 360^\circ$ rotation, extending our prior work. Additionally, the complete kinematic modeling framework incorporates all joints, including the prismatic joint, and demonstrates the coordinated placement of three minor cannulas within the system configuration. The joint-space kinematic model of the proposed tool has been derived and experimentally validated. {\color{black}Overall, all joints separately follow the kinematic model well, with RMSE $<1.5$ mm and MAE $<1$ mm. The combined actuation results also follow the kinematic model closely along the $X$ and $Z$ axes, with RMSE and MAE both  $<1.5$ mm, while the $Y$-axis exhibits a notable deviation from the model.} In future work, we plan to {\color{black}further improve the accuracy by introducing a hysteresis model, implement} the inverse kinematics, and integrate functional end effectors. In addition, the mechanism will be mounted on a 7-DoF robotic manipulator {\color{black}for multi-arm MIS applications}.


\bibliographystyle{IEEEtran}
\bibliography{bibtex}


\end{document}